\documentclass[10pt,twocolumn,letterpaper]{article}

\usepackage{cvpr}
\usepackage{times}
\usepackage{epsfig}
\usepackage{graphicx}
\usepackage{amsmath}
\usepackage{amssymb}
\usepackage{setspace}
\usepackage{cite}
\newtheorem{definition}{Definition}
\usepackage{appendix}
\usepackage[ruled,vlined]{algorithm2e}

\usepackage[breaklinks=true,bookmarks=false]{hyperref}

\cvprfinalcopy 

\def\cvprPaperID{****} 

\begin{document}

\title{Score-CAM: \\
Score-Weighted Visual Explanations for Convolutional Neural Networks}

\author{Haofan Wang, Zifan Wang, Piotr Mardziel\\
Carnegie Mellon University \\
{\tt\small \{haofanw, zifanw\}@andrew.cmu.edu, piotrm@gmail.com}
\and
Mengnan Du, Fan Yang, Xia Hu\\
Texas A\&M University \\
{\tt\small \{dumengnan, nacoyang, xiahu\}@tamu.edu}
\and
Zijian Zhang, Sirui Ding\\
Wuhan University \\
{\tt\small zijianzhang0226@gmail.com, sirui\_ding@whu.edu.cn}
}

\maketitle

\begin{abstract}
   Recently, increasing attention has been drawn to the internal mechanisms of convolutional neural networks, and the reason why the network makes specific decisions. In this paper, we develop a novel post-hoc visual explanation method called Score-CAM based on class activation mapping. Unlike previous class activation mapping based approaches, Score-CAM gets rid of the dependence on gradients by obtaining the weight of each activation map through its forward passing score on target class, the final result is obtained by a linear combination of weights and activation maps. We demonstrate that Score-CAM achieves better visual performance and fairness for interpreting the decision making process. Our approach outperforms previous methods on both recognition and localization tasks, it also passes the sanity check. We also indicate its application as debugging tools. Official code has been released\footnote{https://github.com/haofanwang/Score-CAM}.
\end{abstract}

\section{Introduction}
Explanation of Deep Neural Networks (DNNs) aims to increase the model's transparency to humans so the logic behind the inference can be interpreted in a human understandable way. Among the effort to provide explanations, visualizing a certain quantity of interest, e.g. importance of input features or learned weights, becomes the most straight-forward approach to gain trust from users. Among diverse components in DNNs, the spatial convolution is always the first choice in the feature extraction for both image and language processing. Towards better explanations of convolution operation and Convolutional Neural Network (CNNs), Gradient visualization\cite{simonyan2014very}, Perturbation\cite{ribeiro2016should}, Class Activation Map (CAM) \cite{zhou2016learning} are three of the widely adopted methods.

\begin {figure}[h]
\centering
\includegraphics[width=\columnwidth]{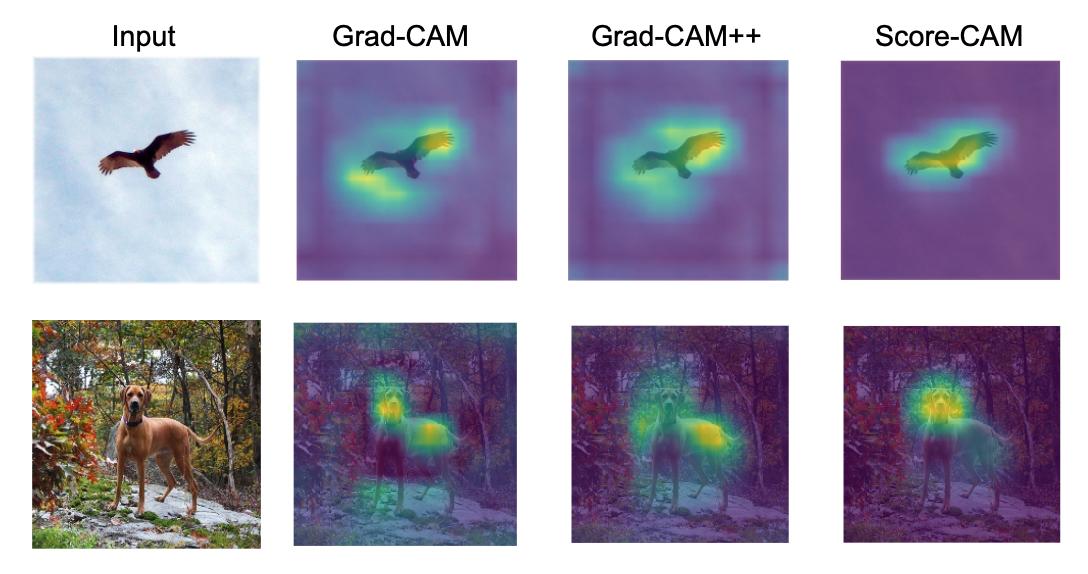}
\caption{Visualization and comparison between our proposed method, Score-CAM with two related work, Grad-CAM and GrdCAM++. Score-CAM shows a potential high concentration at the object in the image.}
\end {figure}

Gradient-based methods backpropagate the gradient of a target class to the input layer to highlight image region that highly influences the prediction. \cite{simonyan2014very} utilizes the derivative of target class score with respect to the input image to generate saliency map. Other works\cite{zeiler2014visualizing,springenberg2014striving,sundararajan2017axiomatic,adebayo2018local,omeiza2019smooth} make additional manipulation on original gradient and visually sharpen the saliency map. These maps are generally of low quality and still noisy\cite{omeiza2019smooth}. Perturbation-Based approaches\cite{ribeiro2016should,petsiuk2018rise,fong2017interpretable, chang2018explaining, dabkowski2017real, wagner2019interpretable} work by perturbing original input and observe the change of the prediction of model. To find minimum region, these approaches usually need additional regularizations\cite{fong2017interpretable} and are time-costing. 

CAM-Based explanation\cite{zhou2016learning,selvaraju2017grad,chattopadhay2018grad} provide visual explanation for a single input with a linear weighted combination of activation maps from convolutional layers. CAM\cite{zhou2016learning} creates localized visual explanations but is architecture-sensitive, a global pooling layer\cite{lin2013network} is required to follow the convolutional layer of interest. Grad-CAM\cite{selvaraju2017grad} and its variations, \eg Grad-CAM++\cite{chattopadhay2018grad}, intend to generalize CAM to models without global pooling layers and finally adopted widely in the community. In this work, we revisit the use of gradient information in GradCAM and discuss our concerns of why gradients may not be an optimal solution to generalize CAM. Further, to address the limitations of gradient-based variations of CAM, we present a new post-hoc visual explanation method, named Score-CAM, where the importance of activation maps are encoded by the global contribution of the corresponding input features instead of the local sensitivity measurement, a.k.a gradient information. We summarize our contributions as follows:

(1) We introduce a novel gradient-free visual explanation method named Score-CAM, which bridges the gap between perturbation-based and CAM-based methods, and represents the weight of activation maps in an intuitively understandable way.

(2) We quantitatively evaluate the fairness of generated saliency maps of Score-CAM on Recognition tasks, specifically Average Drop / Average Increase and Deletion curve / Insertion curve metrics, and show that Score-CAM can find out better evidences of the target class.

(3) We also qualitatively evaluate the visualization and localization performance, and achieve better results on both tasks. Finally, we introduce the effective of its application as a debugging tool to analyze model misbehaviors.

\section{Background}  

Class Activation Mapping (CAM)\cite{zhou2016learning} is a technique for identifying discriminative regions by linearly weighted combination of activation maps of the last convolutional layer before the global pooling layer\footnote{Global Maximum Pooling and Global Average Pooling are two possible implementations but in the original CAM paper it shows the average pooling yields better visual explanations}. To aggregate over multiple channels, CAM incorporates the importance of each channel with the corresponding weight at the following fully connected layer, which generates a score as the class confidence. The biggest restriction of CAM is that not every model is designed with a global pooling layer and even a global pooling layer is present, sometimes more fully connected layers follow before the softmax function, \eg VGG\cite{simonyan2014very}. As a generalization of CAM, Grad-CAM\cite{selvaraju2017grad} is applicable to a broader range of CNN architectures without requiring a specific architecture. Before we dive into the more analysis about CAM and its variations, we first introduce our notations in this paper.
\newline

\noindent \textbf{Notation}
\label{notation}
Consider a CNN neural network $Y = f(X)$ that takes an input $X \in \mathbb{R}^d$ and outputs a probability distribution $Y$, and we denote $Y^c$ as the probablity of being classified as class $c$. For a given layer $l$, let $A_l$ denotes the activations of layer $l$. Specifically, if $l$ is chosen as a convolutional layer, let $A_l^k$ denote the activation map for the $k$-th channel. Also denote the weight of the $k$-th neuron at layer $l$ connecting two layer $l$ and $l+1$ as $w_{l, l+1}$

\begin{definition}[Class Activation Map]
Using the notation in Sec \ref{notation}, consider a model $f$ contains a global pooling layer $l$ that takes the output from the last convolutional layer $l-1$ and feeds the pooled activation to a fully connected layer $l+1$ for classification. For a class of interest $c$, CAM $L^c_{CAM}$ can be defined as
\begin{equation}
    L^c_{CAM} = ReLU(\sum_k \alpha^c_kA^k_{l-1}) 
\end{equation}where \begin{equation}
    \alpha^c_k = w^c_{l, l+1}[k]
\end{equation}$w^c_{l, l+1}[k]$ is the weight for $k$-th neuron after global pooling at layer $l$.
\end{definition} The motivation behind CAM is that each activation map $A^k_l$ contains different spatial information about the input $X$ and the importance of each channel is the weight of the linear combination of the fully connected layer following the global pooling. However, if there is no global pooling layer or there is no (or more than one) fully connected layer(s), CAM will fail due to no definition of $\alpha^c_k$. To resolve the problem, Grad-CAM extends the definition of $\alpha^c_k$ as the gradient of class confidence $Y^c$ w.r.t. the activation map $A_l$. Formally, we have the following definition for Grad-CAM:
\begin{definition}[Grad-CAM]
Using the notation in Sec \ref{notation}, consider a convolutional layer $l$ in a model $f$, given a class of interest $c$, Grad-CAM $L^c_{Grad-CAM}$ can be defined as
\begin{equation}
    L^c_{Grad-CAM} = ReLU(\sum_k \alpha^c_kA^k_l) 
\end{equation}where \begin{equation}
    \alpha^c_k = \text{GP}(\frac{\partial Y^c}{\partial A^k_l})
\end{equation}$\text{GP}(\cdot)$ denoted the global pooling operation \footnote{In the original Grad-CAM paper, the authors use Global Average Pooling and normalize the $\alpha^c_k$ to ensure $\sum_k \alpha^c_k = 1$}.
\end{definition} 

Variations of GradCAM, like GradCAM++\cite{chattopadhay2018grad}, only differentiate in combinations of gradients to represent $\alpha^c_k$. Therefore, we do not explicitly discuss the definitions but will include GradCAM++ as a comparison in the later sections.

Using the gradient to incorporate the importance of each channel towards the class confidence is a natural choice and it also guarantees that Grad-CAM reduces to CAM if there is only one fully connected layer following the chosen layer. Rethinking the concept of ``importance" of each channel in the activation map, we show that Increase of Confidence (definition to follow in Sec \ref{IoC}) is a better way to quantify the channel importance compared to the gradient information. We first discuss some issues regarding the use of gradient to measure importance then we propose our new measurement of channel importance in Sec \ref{sec: method}. 

\subsection{Gradient Issue}
\begin {figure}[!h]
\centering
\includegraphics[width=\columnwidth]{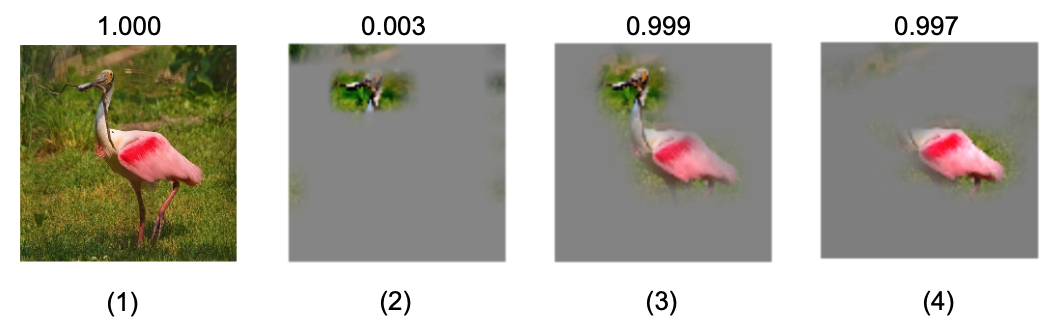}
\caption{(1) is the input image, (2)-(4) are generated by masking input with upsampled activation maps. The weights for activation maps (2)-(4) are 0.035, 0.027, 0.021 respectively. The values above are the increase on target score given (1)-(4) as input. As shown in this example, (2) has the highest weight but cause less increase on target score.}
\label{Figure 1}
\end {figure}
\label{Motivation}
\noindent \textbf{Saturation}
Gradient for a deep neural network can be noisy and also tends to vanish due to the saturation problem for Sigmoid function or the zero-gradient region of ReLU function. One of the consequences is that gradient of the output w.r.t input or the internal layer activation may be noisy visually, which is also one of the issues for Saliency Map method \cite{simonyan2013deep} that attributes input feature importance to the output. An example of gradient can be be visually noisy is shown in Fig \ref{Visualization results}. 

\noindent \textbf{False Confidence}
$L^c_{Grad-CAM}$ is a linear combination of each activation map. Therefore, given two activation map $A^i_l$ and $A^j_l$, if the corresponding weight $\alpha^c_i \geq \alpha^c_j$, we are supposed to claim that the input region which generates $A^i_l$ is at least as important as another region that generates $A^j_l$ towards target class `c'. However, it is easy to find counterexamples with false confidence in Grad-CAM: activation maps with higher weights show lower contribution to the network's output compared to a zero baseline. We randomly select activation maps and upsample them into the input size, then record how much the target score will be if we only keep highlighted region in the activation maps. An example is shown in Fig \ref{Figure 1}. The activation map corresponding to the `head' part receives the highest weight but cause the lowest increase on the target score. This phenomenon may be caused by the global pooling operation on the top of the gradients and the gradient vanishing issue in the network.

\begin{figure*}[h]
\centering
\includegraphics[scale=0.45]{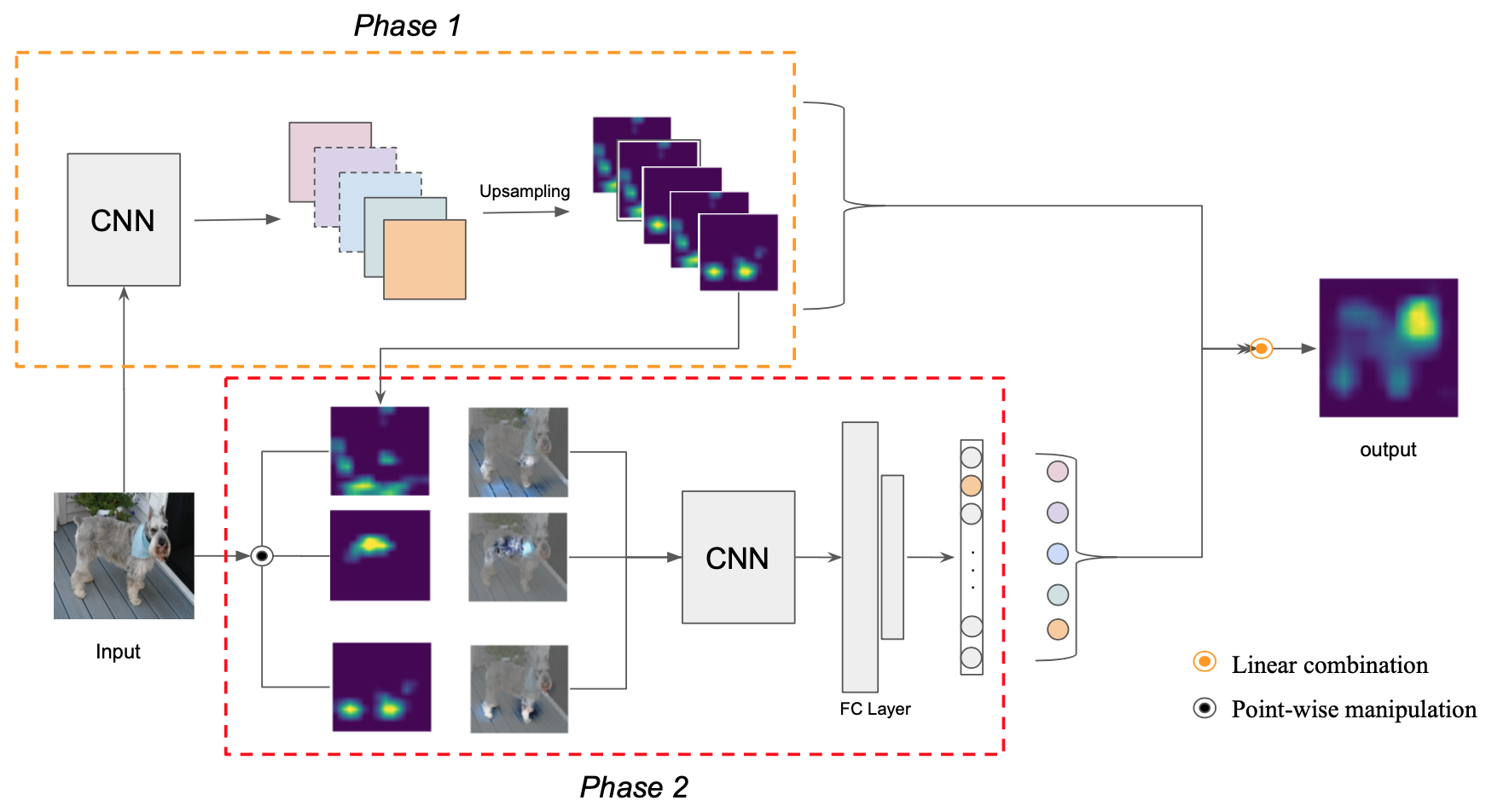}
\caption{Pipeline of our proposed Score-CAM. Activation maps are first extracted in Phase 1. Each activation then works as a mask on original image, and obtain its forward-passing score on the target class. Phase 2 repeats for $N$ times where $N$ is the number of activation maps. Finally, the result can be generated by linear combination of score-based weights and activation maps. Phase 1 and Phase 2 shares a same CNN module as feature extractor.}
\label{Pipeline}
\end{figure*}

\section{Score-CAM: Proposed Approach}
\label{sec: method}
In this section, we first introduce the mechanism of proposed Score-CAM for interpreting CNN-based predictions. The pipeline of the proposed framework is illustrated in Fig \ref{Pipeline}. We first introduce our methodology is then introduced in Sec \ref{Methodology}. Implementation details are followed in Sec \ref{Normalization}.

\subsection{Methodology}
\label{Methodology}
In contrast to previous methods\cite{selvaraju2017grad,chattopadhay2018grad}, which use the gradient information flowing into the last convolutional layer to represent the importance of each activation map, we incorporate the importance as the \emph{Increase of Confidence}.
\begin{definition}[Increase of Confidence]
\label{IoC}
Given a general function $Y=f(X)$ that takes an input vector $X = [x_0, x_1, ..., x_n]^\top$ and outputs a scalar $Y$. For a known baseline input $X_b$, the contribution $c_i$ of $x_i, (i \in [0, n-1])$ towards $Y$ is the change of the output by replacing the $i$-th entry in $X_b$ with $x_i$. Formally,\begin{equation}
    c_i = f(X_b \circ H_i) - f(X_b)
\end{equation}where $H_i$ is a vector with the same shape of $X_b$ but for each entry $h_j$ in $H_i$, $h_j = \mathbb{I}[i = j]$ and $\circ$ denotes Hadamard Product. 
\end{definition}

Some related work has built similar concepts to Def \ref{IoC}. DeepLIFT \cite{shrikumar2017learning} uses the difference of the output given an input compared to the baseline to quantify the importance signals propagating through layers. Two similar concepts \emph{Average Drop \%} and \emph{Increase in Confidence} are proposed by GradCAM++ \cite{chattopadhay2018grad} to evaluate the performance of localization. We generate Def.\ref{IoC} to Channel-wise Increase of Confidence in order to measure the importance of each activation map. 
\begin{definition}
\label{def: cic}
Given a CNN model $Y=f(X)$ that takes an input $X$ and outputs a scalar $Y$. We pick an internal convolutional layer $l$ in $f$ and the corresponding activation as $A$. Denote the $k$-th channel of $A_l$ by $A^k_l$. For a known baseline input $X_b$, the contribution $A^k_l$ towards $Y$ is defined as 
\begin{equation}
    C(A^k_l) = f(X\circ H^k_l) - f(X_b)
\end{equation}
where \begin{equation}
    H^k_l = \text{s(\text{Up($A^k_l$)})}
\end{equation} $\text{Up}(\cdot)$ denotes the operation that upsamples $A^k_l$ into the input size \footnote{We assume that the deep convolutional layer outputs have smaller spatial size compared to the input.} and $s(\cdot)$ is a normalization function that maps each element in the input matrix into [0, 1].
\end{definition}

\noindent \textbf{Use of Upsampling} CIC first upsamples an activation map that corresponds to a specific region in the original input space, and then perturbs the input with the upsampled activation map. The importance of that activation map is obtained by the target score of masked input. Different from \cite{petsiuk2018rise}, where $N$ masks with a  size smaller than image size are generated through Monte Carlo sampling and then upsampled each mask into input size, CIC does not require a process to generate masks. On the contrary, each upsampled activation map not only presents the spatial locations most relevant to an internal activation map, but also can directly work as a mask to perturb the input image. 

\noindent \textbf{Smoothing with Normalization} 
\emph{Increase of Confidence} essentially creates a binary mask $H_i$ on the top of the input with only the feature of interest retained in the input. However, the binary mask may not be a reasonable choice when we are not interested in one pixel but a specific region in the input image. Instead of setting all elements to binary values, in order to a generate smoother mask $H^k_L$ for an activation map, we normalize the raw activation values in each activation map into $[0,1]$. We use the following normalization function in the Algorithm 1 of Score-CAM:
\begin{equation}
s(A^k_{l}) = \frac{A^k_{l} - \min \limits_{} A_{l}^k}{\max \limits_{} A_{l}^k - \min \limits_{} A_{l}^k}
\end{equation}
Finally, we describe our proposed visual explanation method Score-CAM in Def.\ref{Score-CAM definition}. The complete detail of the implementation is described in Algorithm 1. 

\begin{figure*}[h]
\centering
\includegraphics[scale=0.18]{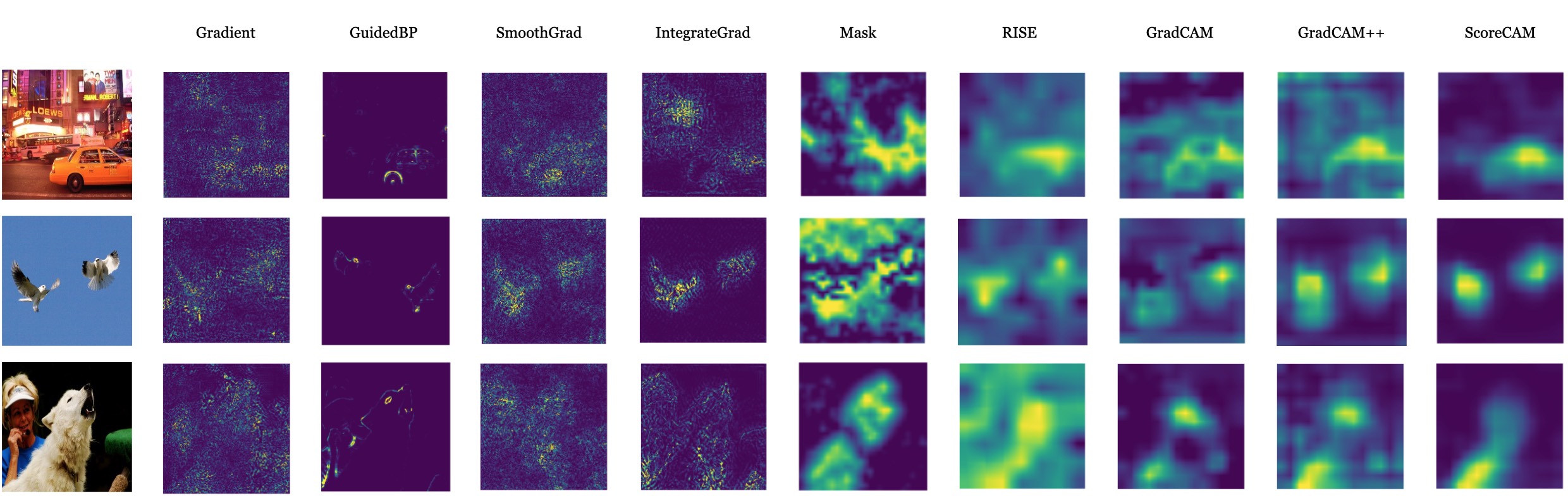}
\caption{Visualization results of Vanilla  Backpropagation\cite{simonyan2013deep}, Guided Backpropagation\cite{springenberg2014striving}, SmoothGrad\cite{smilkov2017smoothgrad}, IntegrateGrad\cite{sundararajan2017axiomatic}, Mask\cite{fong2017interpretable}, RISE\cite{petsiuk2018rise}, Grad-CAM\cite{selvaraju2017grad}, Grad-CAM++\cite{chattopadhay2018grad} and our proposed Score-CAM. More results are provided in Appendix.}
\label{Visualization results}
\end{figure*}

\begin{definition}[Score-CAM]
\label{Score-CAM definition}
Using the notation in Sec \ref{notation}, consider a convolutional layer $l$ in a model $f$, given a class of interest $c$, Score-CAM $L^c_{Score-CAM}$ can be defined as
\begin{equation}
    L^c_{Score-CAM} = ReLU(\sum_k \alpha^c_kA^k_l) 
\end{equation} where \begin{equation}
    \alpha^c_k = C(A^k_l)
\end{equation}where $C(\cdot)$ denotes the CIC score for activation map $A^k_l$.
\end{definition}

\begin{algorithm}[h]
\SetAlgoLined
\KwIn{Image $X_0$, Baseline Image $X_b$, Model $f(X)$, class $c$, layer $l$}
\KwOut{$L^c_{Score-CAM}$}
 initialization\;
 // get activation of layer $l$\;
 $M \leftarrow []$, $A_l \leftarrow f_l(X)$\\
 $C\leftarrow$ the number of channels in $A_l$\\
 \For{k in $[0, ..., C-1]$}{
  $M^k_l \leftarrow$ Upsample$(A^k_l)$\\
  // normalize the activation map\;
  $M^k_l \leftarrow$ s$(M^k_l)$\\
  // Hadamard product\;
  $M$.append($M^k_l \circ X_0$)
 }
 $M \leftarrow $ Batchify($M$)\\
 //$f^c(\cdot)$ as the logit of class $c$\;
 $S^c \leftarrow f^c(M) - f^c(X_b)$\\
 // ensure $\sum_k\alpha^c_k=1$ in the implementation\;
 $\alpha^c_k \leftarrow \frac{\exp(S^c_k)}{\sum_k \exp(S^c_k)}$\\
 $L^c_{Score-CAM} \leftarrow ReLU(\sum_k \alpha^c_kA^k_l)$
 \caption{Score-CAM algorithm}
\end{algorithm}

Similar to \cite{selvaraju2017grad,chattopadhay2018grad}, we also apply a ReLU to the linear combination of maps because we are only interested in the features that have a positive influence on the class of interest. Since the weights come from the CIC score corresponding to the activation maps on target class, Score-CAM gets rid of the dependence on gradient. Although the last convolution layer is a more preferable choice because it is end point of feature extraction \cite{selvaraju2017grad}, any intermediate convolutional layer can be chosen in our framework.

\subsection{Normalization on Score}
\label{Normalization}
Each forward passing in neural network is independent, the score amplitude of each forward propagation is unpredictable and not fixed. The relative output value (post-softmax) after normalization is more reasonable to measure the relevance than absolute output value (pre-softmax). Thus, in Score-CAM, we represent weight as post-softmax value, so that the score can be rescaled into a fixed range.

\begin {figure}[h]
\centering
\includegraphics[scale=0.3]{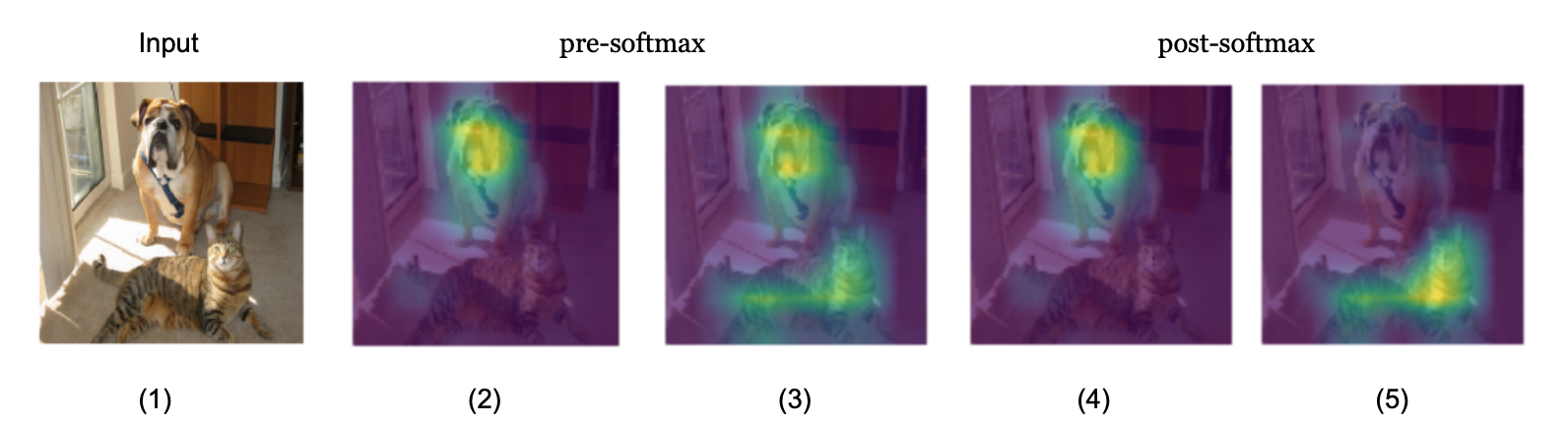}
\caption{Effect of normalization. (2) and (4) are w.r.t `boxer dog’, (3) and (5) are w.r.t `tiger cat’. As shown, pre-softmax and post-softmax show a difference on class discrimination ability. We adopt post-softmax value in all other section in this paper.}
\label{Effect of normalization}
\end {figure}

Because of the varied range in each prediction, whether or not using softmax makes a difference. An interesting discovery is shown in Fig \ref{Effect of normalization}. The model predicts the input image as `dog' which can be correctly highlighted no matter which type of score is adopted. But for target class `cat', Score-CAM highlight both region of `dog' and `cat' if using pre-softmax logit as weight. On the contrary, Score-CAM with softmax can well distinguish two different categories, even though the prediction probability of `cat' is lower than the probability of `dog'. Normalization operation equips Score-CAM with good class discrimination ability.

\section{Experiments}
In this section, we conduct experiments to evaluate the effectiveness of the proposed explanation method. First, we qualitatively evaluate our approach via visualization on ImageNet in Sec \ref{Evaluating Visualization}. Second, we evaluate the fairness of explanation (how importance the highlighted region is for model decision) on image recognition in Sec \ref{Evaluating Faithfulness}. In Sec \ref{Evaluating Localization} we show the effectiveness for class-conditional localization of objects in a given image. The sanity check is followed in Sec \ref{sanity check}. Finally, we employ Score-CAM as a debugging tool to analyze model misbehaviors in Sec \ref{Harnessing Explanations For Model Analysis}.

In the following experiments, unless stated otherwise, we use pre-trained VGG16 network from the Pytorch model zoo\footnote{https://github.com/pytorch/vision} as a base model, more visualization results on other network architectures are provided in Appendix. Publicly available object classification dataset, namely, ILSVRC2012 val \cite{russakovsky2015imagenet} is used in our experiment. For the input images, we resize them to ($224$ $\times$ $224$ $\times$ $3$), transform them to the range $[0, 1]$, and then normalize them using mean vector $[0.485, 0.456, 0.406]$ and standard deviation vector $[0.229, 0.224, 0.225]$. No further pre-processing is performed.

\begin{table*}[htbp]
\centering 
\setlength{\tabcolsep}{1.5mm}{
\caption{Evaluation results on Recognition (lower is better in Average Drop, higher is better in Average Increase).}
\begin{tabular}{lllllllllllll}
\hline
 Method  &  Mask & RISE  & GradCAM & GradCAM++ & ScoreCAM \\
\hline
Average Drop(\%) & 63.5 & 47.0 & 47.8 & 45.5 & \textbf{31.5} \\
\hline
Average Increase(\%) & 5.29 & 14.0 & 19.6 & 18.9 & \textbf{30.6} \\
\hline
\label{table2}
\end{tabular}}
\end{table*}

\subsection{Qualitative Evaluation via Visualization}
\label{Evaluating Visualization}
\subsubsection{Class Discriminative Visualization}
We qualitatively compare the saliency maps produced by 8 state-of-the-art methods, namely gradient-based, perturbation-based and CAM-based methods. Our method generates more visually interpretable saliency maps with less random noises. Results are shown in Fig \ref{Visualization results}, more examples are provided in Appendix. As shown, in Score-CAM, random noises are much less than Mask\cite{fong2017interpretable}, RISE\cite{petsiuk2018rise}, Grad-CAM\cite{selvaraju2017grad} and Grad-CAM++\cite{chattopadhay2018grad}. Our approach can also generate smoother saliency maps comparing with gradient-based methods.

\begin {figure}[h]
\centering
\includegraphics[width=\columnwidth]{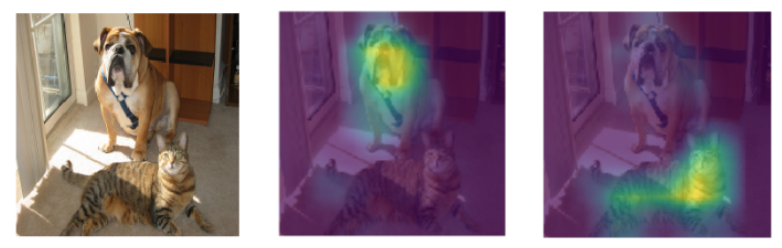}
\caption{Class discriminative result. The middle plot is generated w.r.t `bull mastiff', and the right plot is generated w.r.t `tiger cat'.}
\label{Class discriminative result.}
\end {figure}

We demonstrate that Score-CAM can distinguish different classes as shown in Fig \ref{Class discriminative result.}. The VGG-16 model classifies the input as `bull mastiff' with $49.6\%$ confidence and `tiger cat' with $0.2\%$ confidence, our model correctly gives the explanation locations for both of two categories, even though the prediction probability of the latter is much lower than the probability of the former. It is reasonable to expect Score-CAM to distinguish different categories, because the weight of each activation map is correlated with the response on target class, and this equips Score-CAM with good class discriminative ability. 

\begin {figure}[h]
\centering
\includegraphics[width=\columnwidth]{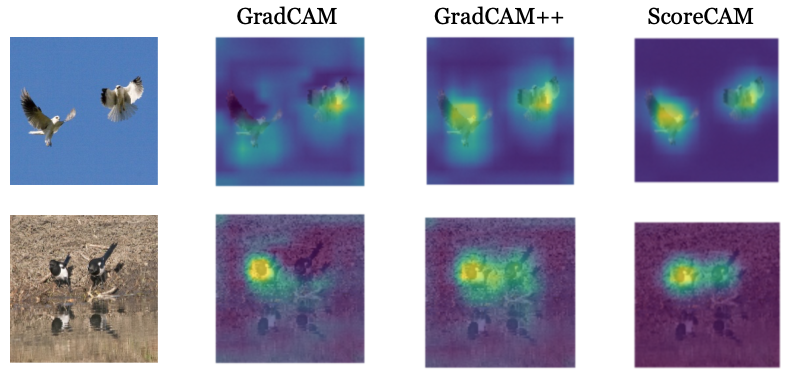}
\caption{Results on multiple objects. As shown in this example, Grad-CAM only tends to focus on one object, while Grad-CAM++ can highlight all objects. Score-CAM further improves the quality of finding all evidences.}
\label{Results on multiple objects}
\end {figure}

\subsubsection{Multi-Target Visualization}
Score-CAM can not only locate single object accurately, but also show better performance on locating multiple same class objects than previous works. The result is shown in Fig \ref{Results on multiple objects}, Grad-CAM\cite{selvaraju2017grad} tends to only capture one object in the image, Grad-CAM++\cite{chattopadhay2018grad} and Score-CAM both show ability to locate multiple objects, but the saliency maps of Score-CAM are more focused than Grad-CAM++. 

As the weight of each activation map is represented by its score on the target class, each target object with a high confidence score predicted by the model can be highlighted independently. Therefore, all evidences related to target class can get responses and are assembled through linear combination.

\begin{table*}[h]
\centering 
\setlength{\tabcolsep}{1.5mm}{
\caption{Comparative evaluation on Energy-Based Pointing Game (higher is better).}
\begin{tabular}{lllllllllllll}
\hline
& Grad & Smooth  & Integrated  & Mask & RISE  & GradCAM  & GradCAM++  & ScoreCAM
\\
\hline
Proportion(\%) & 41.3 & 42.4 & 44.7 & 56.1 & 36.3 & 48.1 & 49.3 & \textbf{63.7} \\
\hline
\label{table1}
\end{tabular}}
\end{table*}

\subsection{Faithfulness Evaluation via Image Recognition}
\label{Evaluating Faithfulness}
We first evaluate the faithfulness of the explanations generated by Score-CAM on the object recognition task as adopted in \cite{chattopadhay2018grad}. The original input is masked by point-wise multiplication with the saliency maps to observe the score change on the target class. In this experiment, rather than do point-wise multiplication with the original generated saliency map, we slightly modify by limiting the number of positive pixels in the saliency map (50\% of pixels of the image are muted in our experiment). We follow the metrics used in \cite{chattopadhay2018grad} to measure the quality, the Average Drop is expressed as $\sum^N_{i=1}\frac{\max(0,Y^c_i - O^c_i)}{Y^c_i}\times100$, the Increase In Confidence (also denote as Average Increase) is expressed as $\sum^N_{i=1}\frac{Sign(Y^c_i < O^c_i)}{N}$, where $Y^c_i$ is the predicated score for class $c$ on image $i$ and $O^c_i$ is the predicated score for class $c$ with the explanation map region as input. $Sign$ presents an indicator function that returns $1$ if input is True. Experiment conducts on the ImageNet (ILSVRC2012) validation set, 2000 images are randomly selected. Result is reported in Table \ref{table2}.

As shown in Table \ref{table2}, Score-CAM achieves 31.5\% average drop and 30.6\% average increase respectively, and outperforms other perturbation-based and CAM-based methods by large scale. A good performance on recognition task reveals that Score-CAM can successfully find out the most distinguishable region of the target object, rather than just finding what human think is important. Here we do not compare with gradient-based methods, because of their different visual properties. Results on recognition task demonstrates that Score-CAM could more faithfully reveal the decision making process of the original CNN model than previous approaches.

\begin{figure*}[h]
\centering
\includegraphics[scale=0.35]{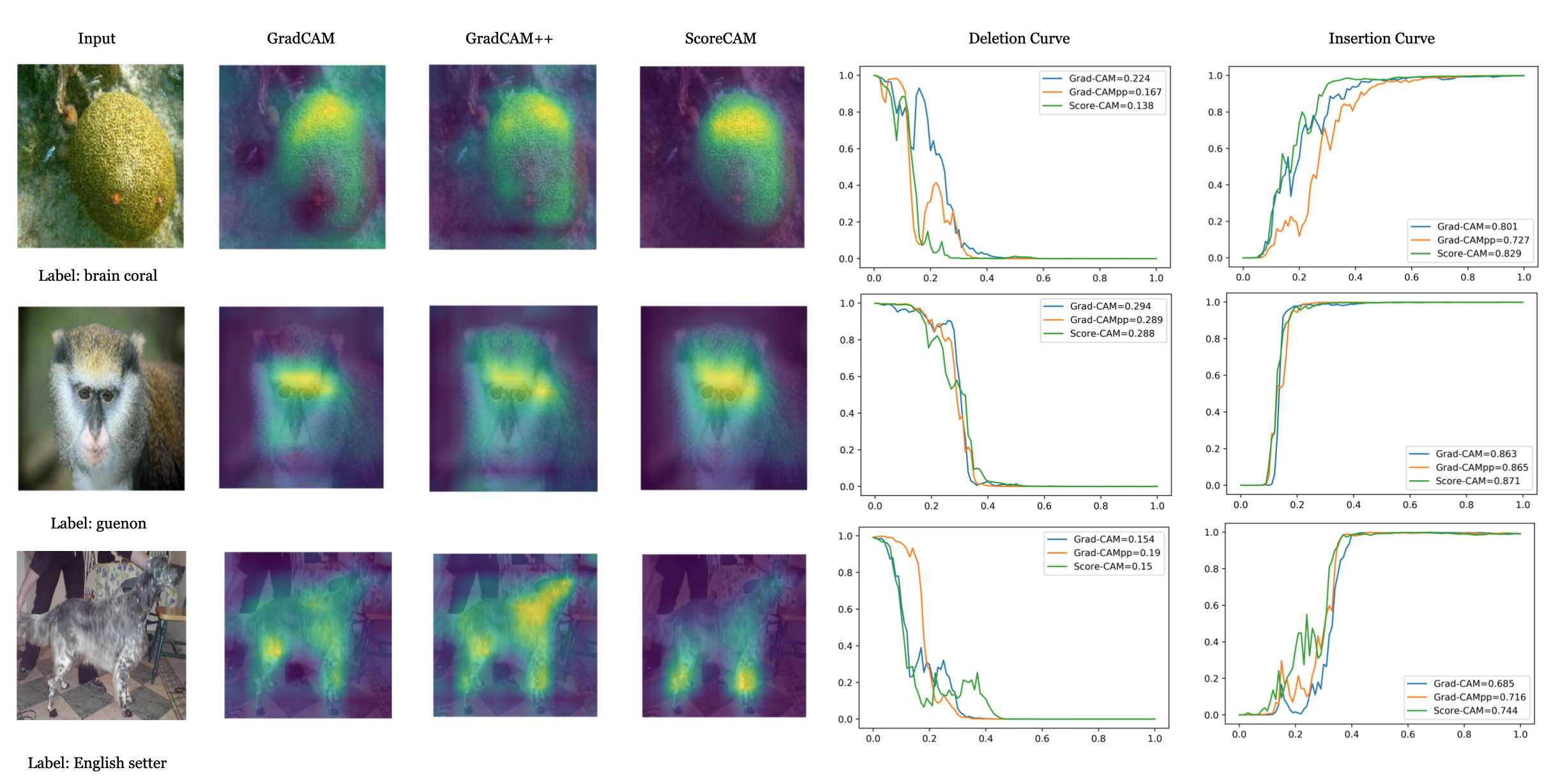}
\caption{Grad-CAM, Grad-CAM++ and Score-CAM generated saliency maps for representative images with deletion and insertion curves. In deletion curve, a better explanation is expected to drop faster, the AUC should be small, while in increase curve, it is expected to increase faster, the AUC should be large.}
\label{AUC}
\end{figure*}

Furthermore, for a more comprehensive comparison, we also evaluate our method on deletion and insertion metrics which are proposed in \cite{petsiuk2018rise}. As supplementary to Average Drop and Average Increase metrics, the deletion metric measures a decrease in the probability of the predicted class as more and more important pixels are removed, where the importance of each pixel is obtained from the generated saliency map. A sharp drop and thus a low area under the probability curve (as a function of the fraction of removed pixels) means a good explanation. The insertion metric, on the other hand, measures the increase in probability as more and more pixels are introduced, with higher AUC indicative of a better explanation. 

\begin{table}[h]
\centering 
\setlength{\tabcolsep}{1.5mm}{
\caption{Comparative evaluation in terms of deletion (lower is better) and insertion (higher is better) scores.}
\begin{tabular}{lllllllllllll}
\hline
& Grad-CAM & Grad-CAM++ & Score-CAM
\\
\hline
Insertion & 0.357 & 0.346 & \textbf{0.386}\\
\hline
Deletion & 0.089  & 0.082 & \textbf{0.077} \\ 
\hline
\label{table3}
\end{tabular}}
\end{table}

There are several ways of removing pixels from an image\cite{dabkowski2017real}, all of these approaches have different pros and cons. Thus, in this experiment, we simply remove or introduce pixels from an image by setting the pixel values to zero or one with step 0.01 (remove or introduce 1\% pixels of the whole image each step). Example are shown in Fig \ref{AUC}. The average result over 2000 images is reported in Table \ref{table3}, where our approach achieves better performance on both metrics compared with gradient-based CAM methods.

\subsection{Localization Evaluation}
\label{Evaluating Localization}
In this section, we measure the quality of the generated saliency map through localization ability. Extending from pointing game which extracts maximum point in saliency map to see whether the maximum falls into object bounding box, we treat this problem in an energy-based perspective. Instead of using only the maximum point, we care about how much energy of the saliency map falls into the target object bounding box. Specifically, we first binarize the input image with the bounding box of the target category, the inside region is assigned to 1 and the outside region is assigned to 0. Then, we point-wise multiply it with generated saliency map, and sum over to gain how much energy in target bounding box. We denote this metric as $Proportion = \frac{\sum{L^c_{(i,j)\in bbox}}}{\sum{L^c_{(i,j)\in bbox}} + \sum{L^c_{(i,j)\notin bbox}}}$, and call this metric an energy-based pointing game. 

As we observe, it is common in the ILSVRC validation set that the object occupies most of the image region, which makes these images not suitable for measure the localization ability of the generated saliency maps. Therefore, we randomly select images from the validation set by removing images where object occupies more than 50\% of the whole image, for convenience, we only consider these images with only one bounding box for target class. We experiment on 500 random selected images from the ILSVRC 2012 validation set. Evaluation result is reported in Table \ref{table1}, which shows that our method outperforms previous works by a large scale, more than 60\% energy of saliency map falls into the ground truth bounding box of the target object. This is also a corroboration that the saliency map generated by Score-CAM comes with less noises. We don't compare with Guided BackProp\cite{springenberg2014striving} because it works similar to an edge detector rather than saliency map (heatmap). In addition, it should be more accurate to evaluate on segmentation label rather than object bounding box, we will add it in our future work.

\begin{figure}[h]
\centering
\includegraphics[scale=0.32]{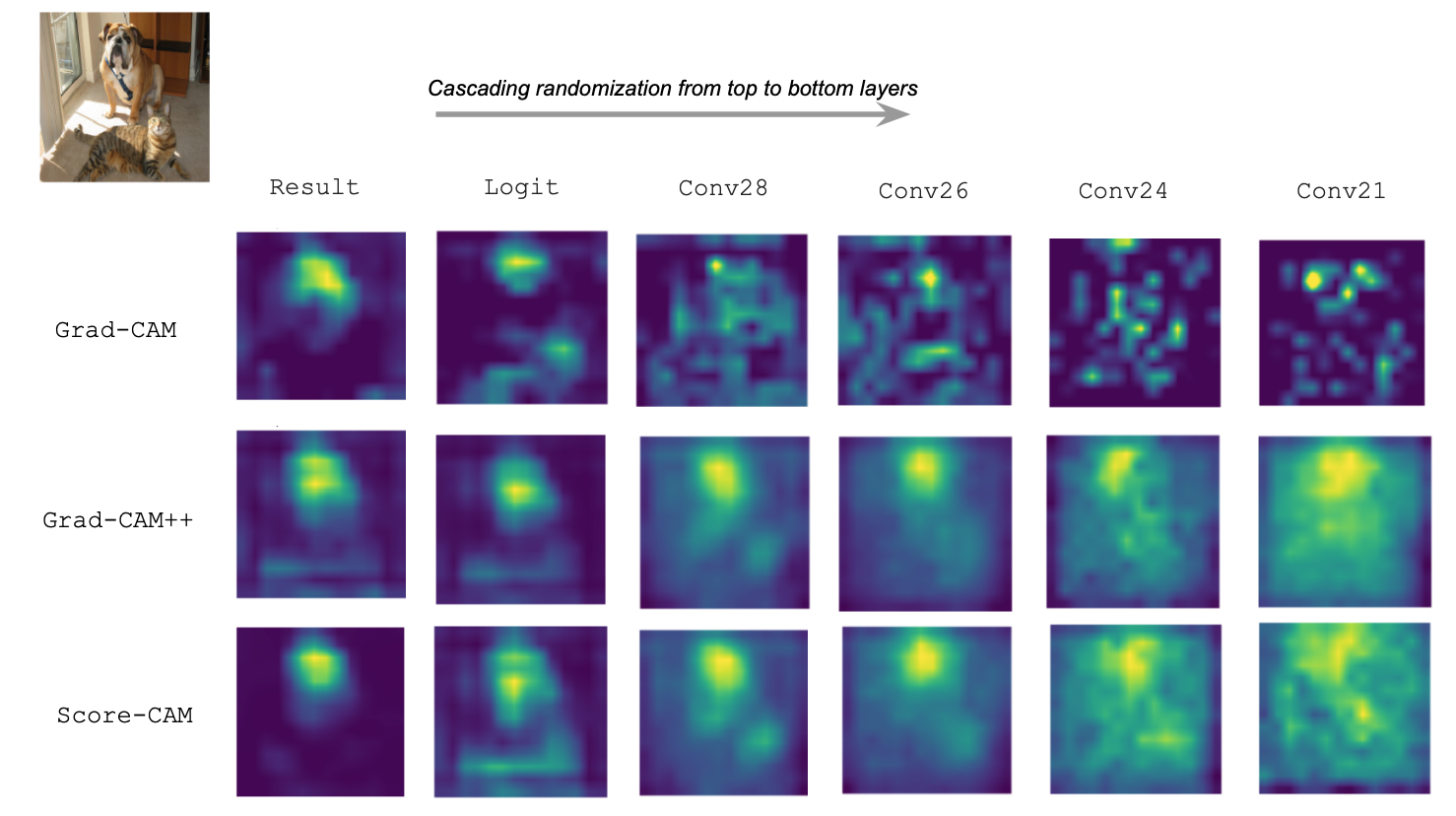}
\caption{Sanity check results by randomization. The first column is the original generated saliency maps. The following columns are results after randomizing from top the layers respectively. The results show sensitivity to model parameters, the quality of saliency maps can reflect the quality of the model. All three types of CAM pass the sanity check.}
\label{sanity}
\end{figure}

\subsection{Sanity Check}
\label{sanity check}
\cite{adebayo2018sanity} finds that reliance, solely, on visual assessment can be misleading. Some saliency methods\cite{springenberg2014striving} are independent both of the model and of the data generating process. We adopt model parameter randomization test proposed in \cite{adebayo2018sanity}, to compare the output of Score-CAM on a trained model with the output of a randomly initialized untrained network of the same architecture. As shown in Fig \ref{sanity}, as the same as Grad-CAM and Grad-CAM++, Score-CAM also passes the sanity check. The Score-CAM result is sensitive to model parameter and therefore can reflect the quality of model.

\subsection{Applications}
\label{Harnessing Explanations For Model Analysis}
A good post-hoc visual explanation should not only tell where does the model look at, but also help researchers analyze their models. We claim that much previous work treat visual explanation as a way to do localization, but ignore the usefulness in helping to analyze the original model. In this part, we suggest how to harness the explanations generated by Score-CAM for model analysis, and provide insights for future exploration.

\begin {figure}[h]
\centering
\includegraphics[width=\columnwidth]{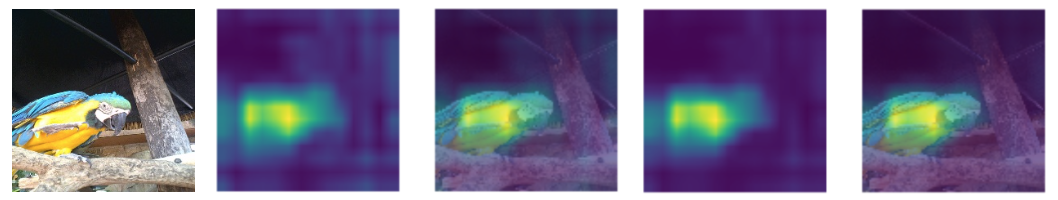}
\caption{The left is generated by no-finetuning VGG16 with 22.0\% classification accuracy , the right is generated by finetuning VGG16 with 90.1\% classification accuracy. It shows that the saliency map becomes more focused as the increasing of classification accuracy.}
\end {figure}

We observe that Score-CAM can work well on localization task even the classification performance of the model is bad, but as the classification performance improves, the noise in saliency map decreases and focuses more on important region. The noise suggests the classification performance. This also can work as a hint to determine whether a model has converged, if the generated saliency map does not change anymore, the model may have converged.

\begin {figure}[h]
\centering
\includegraphics[width=\columnwidth]{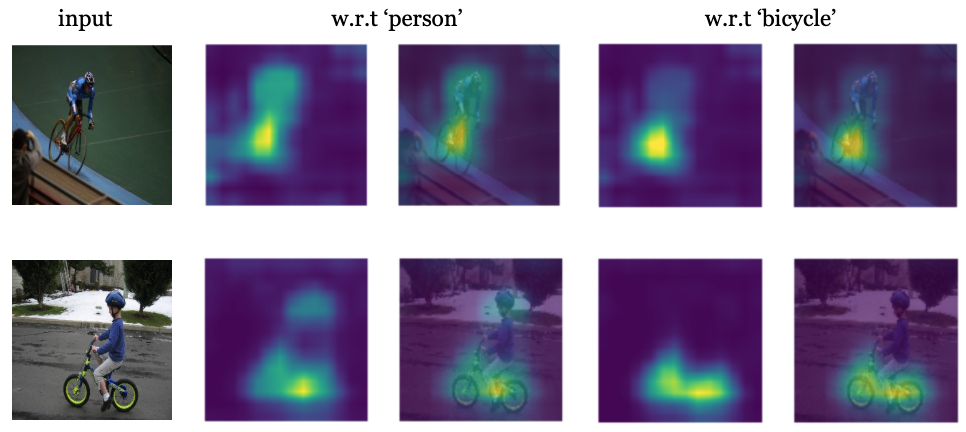}
\caption{The left column is input example, middle is saliency map w.r.t predicted class (person), right is saliency map w.r.t target class(bicycle).}
\label{why the model makes a wrong prediction}
\end {figure}

Besides, Score-CAM can help diagnose why the model makes a wrong prediction and identify dataset bias. The image with label `bicycle' is classified as `person' in Fig \ref{why the model makes a wrong prediction}. Saliency maps for both classes are generated. By comparing the difference, we know that `person' is correlated with `bicycle' because `person' appears in most of `bicycle' images in training set, and `person' region is the most distractive part that leads to mis-classification.

\section{Conclusion}
In this paper, we proposed a new kind of CAM variants, named Score-CAM method, as a better visual explanation. Score-CAM incorporates Increase in Confidence in the designing of weight for each activation map, gets rid of the dependence on gradients and has a more reasonable weight representation. We provide an in-depth analysis of motivation, implementation, qualitative and quantitative evaluations. Our method outperforms all previous CAM-based methods and other state-of-the-art methods in recognition and localization evaluation metrics. Future work involves explore the connection of different weight representations in other kind of CAM variants.

{\small
\bibliographystyle{ieee}
\bibliography{paper}
}

\begin{appendix}
\onecolumn 

\section*{\centering\textcolor{black}{Appendix}}

\begin{figure*}[h]
\centering
\includegraphics[width=\columnwidth]{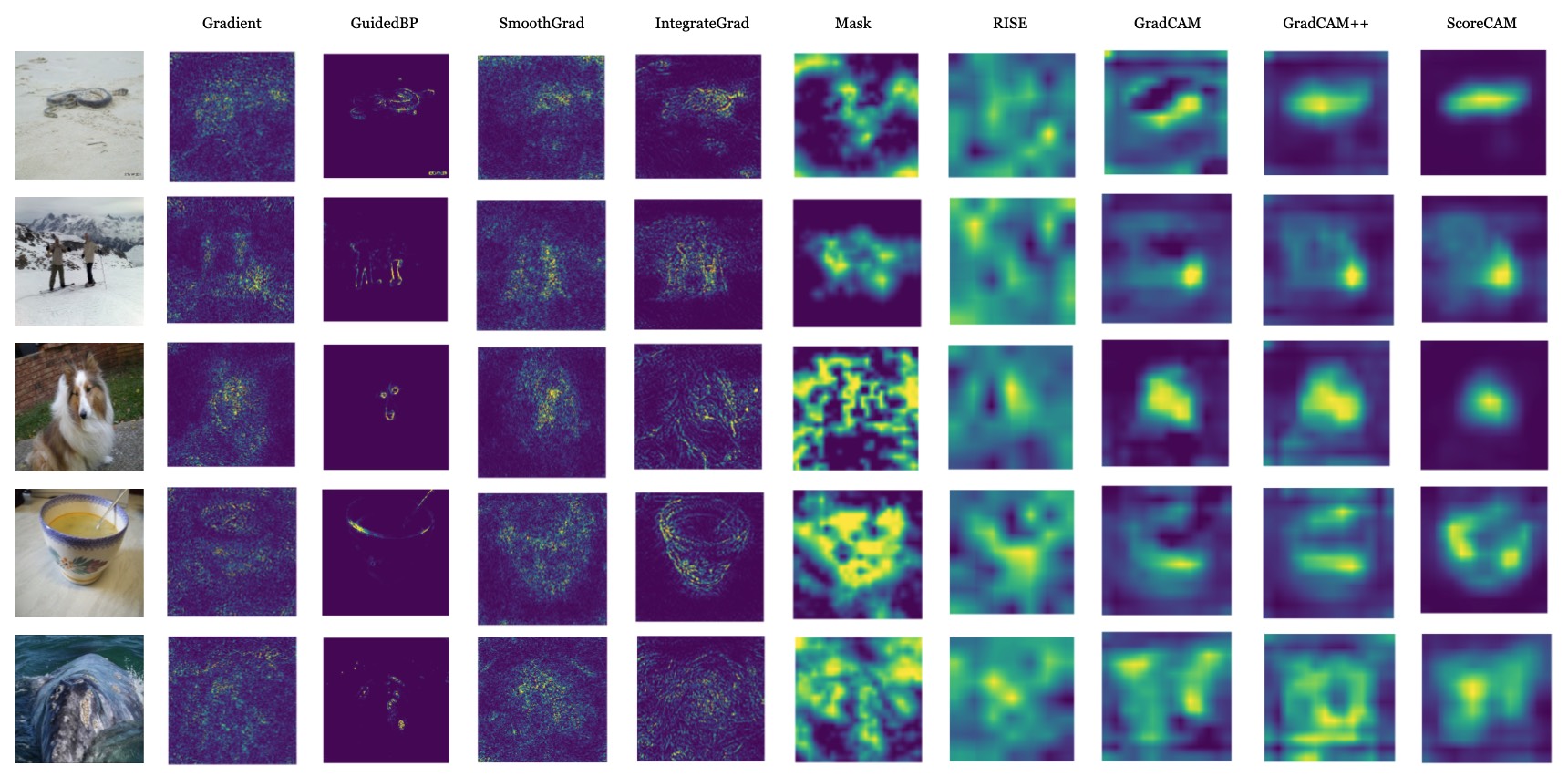}
\includegraphics[width=\columnwidth]{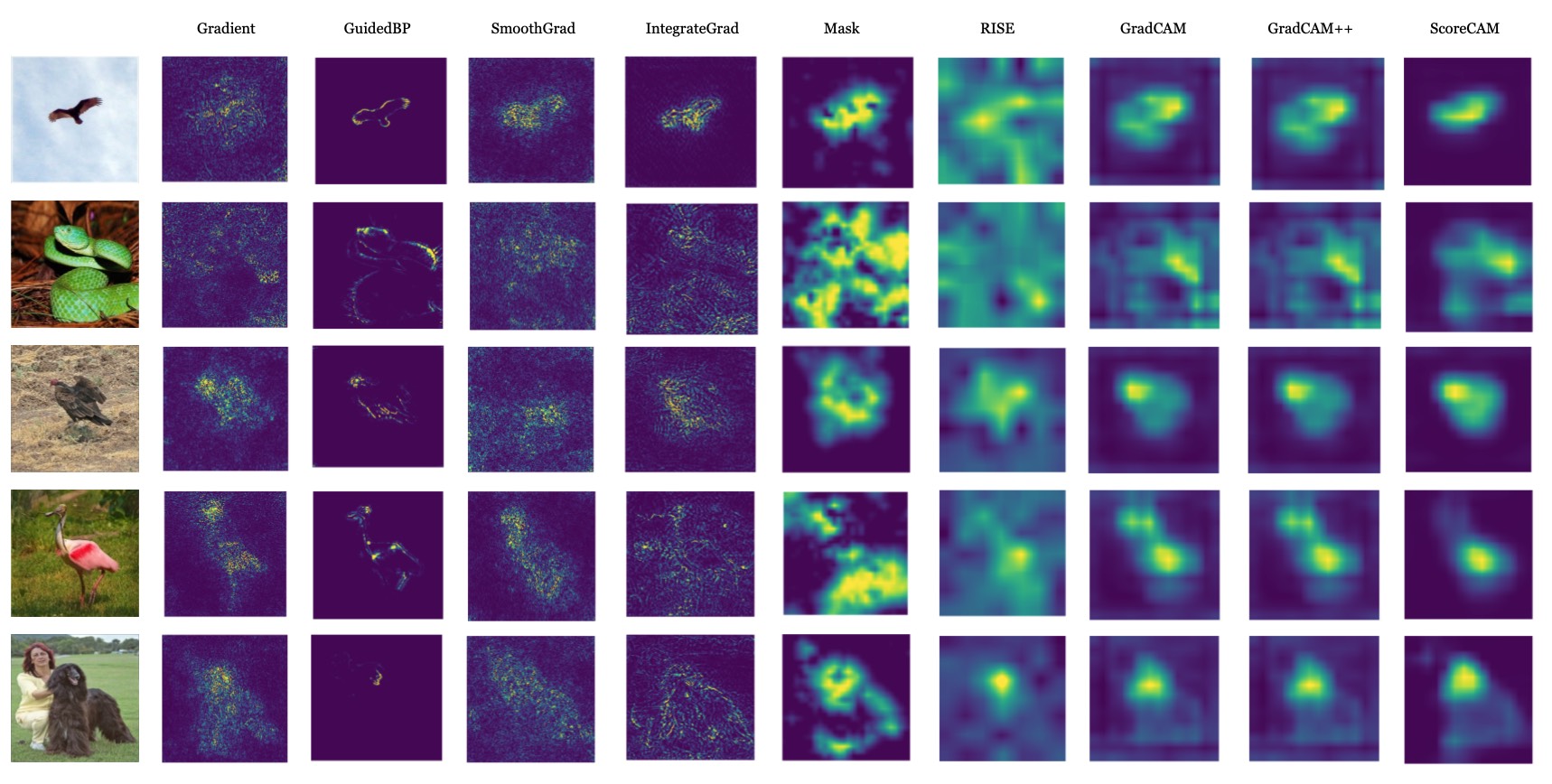}
\caption{Visualization results on single object.}
\end{figure*}

\begin{figure*}[h]
\centering
\includegraphics[width=\columnwidth]{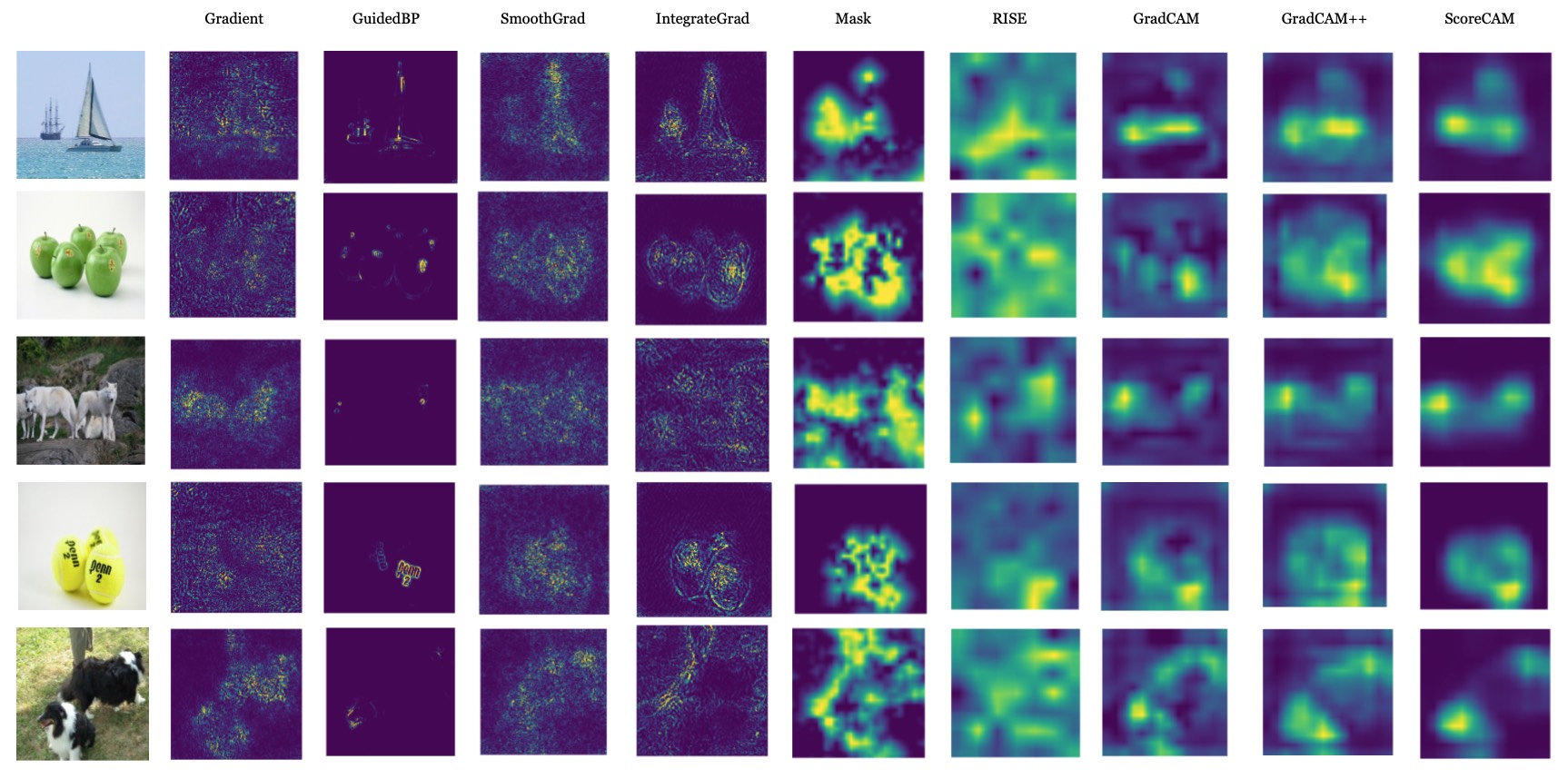}
\caption{Visualization results on multiple objects.}
\end{figure*}

\begin{figure*}[h]
\centering
\includegraphics[width=\columnwidth]{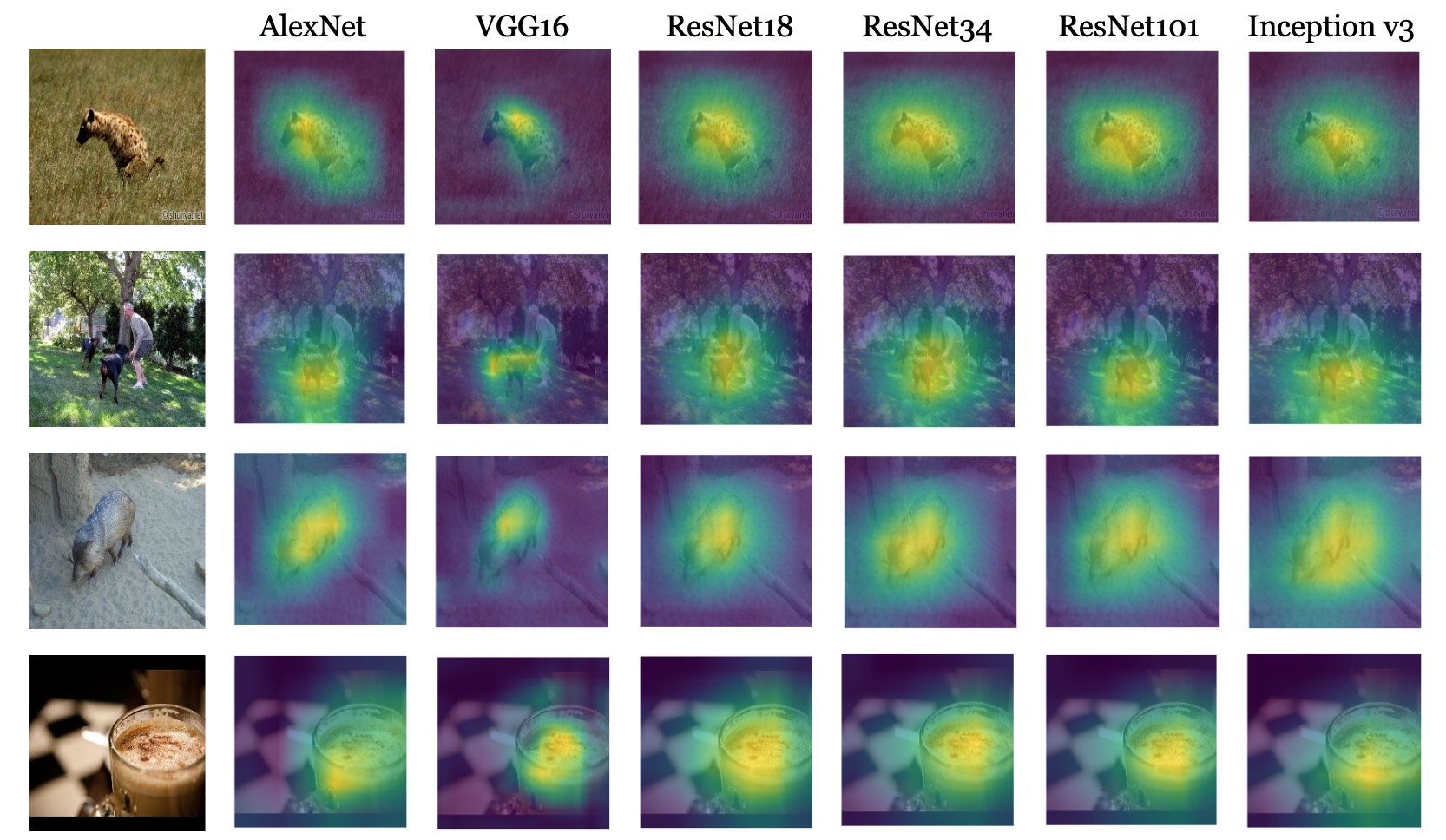}
\caption{Score-CAM results on other model architectures.}
\label{other models}
\end{figure*}
\end{appendix}

\end{document}